\documentclass{article}
\usepackage{spconf,amsmath,graphicx}
\usepackage{caption}
\usepackage{subcaption}
\usepackage{eucal}
\usepackage{multirow}
\usepackage{amsfonts}

\title{EffMulti: Efficiently Modeling Complex Multimodal Interactions for Emotion Analysis}
\twoauthors
 {Feng Qiu,
     Chengyang Xie,
     Yu Ding$^\ast$\thanks{* Corresponding author.}}
	{Virtual Human Group, \\Netease Fuxi AI Lab, China}
 {Wanzeng Kong}
	{College of Computer Science,\\ Hangzhou Dianzi University}
\begin{document}
%
\maketitle
\begin{abstract}
Humans are skilled in reading the interlocutor’s emotion from multimodal signals, including spoken words, simultaneous speech, and facial expressions. It is still a challenge to effectively decode emotions from the complex interactions of multimodal signals. In this paper, we design three kinds of multimodal latent representations to refine the emotion analysis process and capture complex multimodal interactions from different views, including a intact three-modal integrating representation, a modality-shared representation, and three modality-individual representations. Then, a modality-semantic hierarchical fusion is proposed to reasonably incorporate these representations into a comprehensive interaction representation. The experimental results demonstrate that our EffMulti outperforms the state-of-the-art methods. The compelling performance benefits from its well-designed framework with ease of implementation, lower computing complexity, and less trainable parameters.
\end{abstract}

\section{Introduction}

\maketitle

Endowing machines with emotional intelligence has been a long-standing goal for engineers and researchers working on artificial intelligence \cite{KDD1}. It contributes to technology-mediated interaction, such as interactive virtual reality (VR) games, virtual tutors and public opinion analysis. Humans are very skilled in conveying and understanding emotions through multimodal signals, which are supplementary and complementary to each other by skillfully interacting with each other \cite{zadeh2017tensor}.

It has still been a challenging task to effectively explore complex multimodal interactions for emotion analysis. The exploration aims at learning a fusion representation to offer effective information for emotion analysis. Previous works \cite{zadeh2017tensor, MFN, LMF} are dedicated to the straightforward learning of a fusion representation by integrating multimodal original features. These methods are often challenged by the modality gaps persisting between heterogeneous modalities \cite{MISA}. So these methods may be incapable of fully modeling the complex interaction of multiple modalities.

Later on, another work MISA \cite{MISA} refines the learning of fusion representation by decoupling all three modal features into a modality-invariant and -variant representation and then fusing them. It approves the benefit of the decoupling operation in minimizing redundancy between multimodal information and in modeling more complex multimodal interaction. On the other hand, the decoupling operation in MISA requires several auxiliary orthogonal constraint losses and hyper-parameters balancing these losses, as well as a large quantity of additional trainable parameters. These requirements make the training complex. 

To investigate complex multimodal interactions, our work EffMulti takes into account three types of multimodal interaction. These representations encode and shape the multimodal interactions from different views:
(1) The modality-full interaction representation reserves intact information of the single modality for emotion analysis; 
(2) The modality-shared interaction representation focuses on the contribution of coordination among multiple modalities; 
(3) The modality-specific interaction representations highlight the contribution of individual specific information excluding common information. 

Then, these three types of interaction representations are merged into a comprehensive fusion representation to unleash the full representation capacity of the multimodal interaction for emotion analysis. Importantly, our efforts are also made on creating an ease-training, low computing cost but effective framework for modeling the above multimodal complex interactions.

Our contributions can be summarised as follows: (1) This work emphasizes and validates the importance of modeling multi-view multimodal interactions for emotion analysis with better performance than the state-of-the-art works. (2) This work intentionally develops a simple but effective, ease-training, fast convergence, lower computation cost and less learnable parameters framework dedicated for multimodal sentiment analysis.

\begin{figure*}[ht]
\centering
\includegraphics[width=0.95\textwidth]{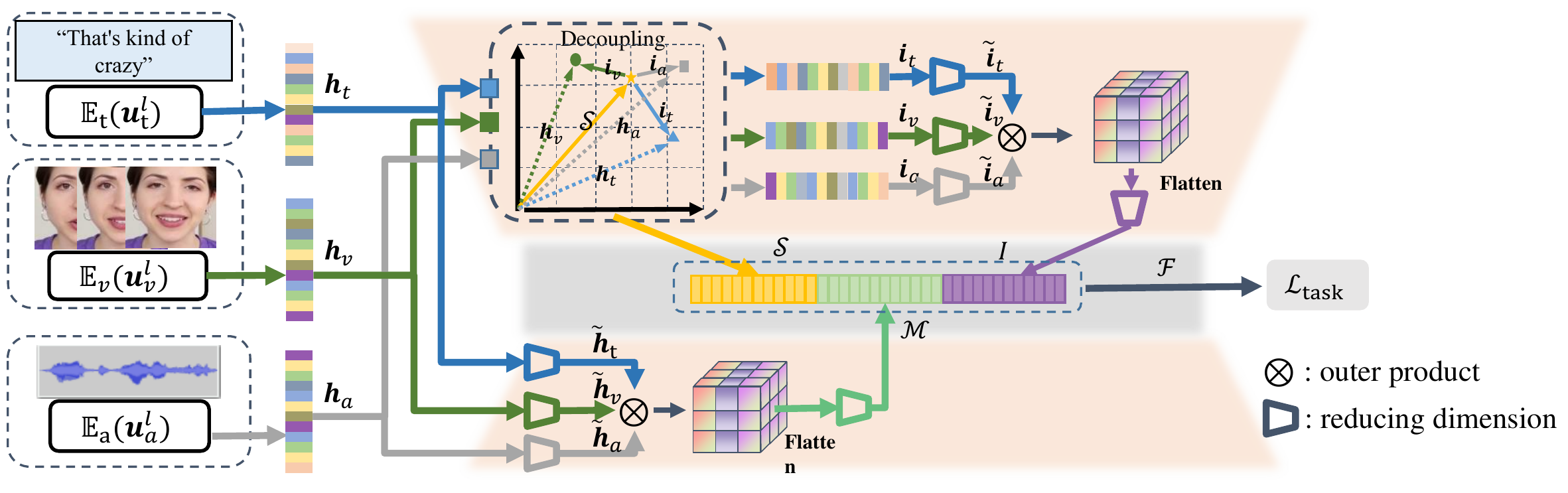}
\caption{Illustration of EffMulti framework. The \emph{Unimodal Encoding Module} takes the frame-level or utterance-level unimodal features as input and produces utterance-based unimodal representations (${\textbf{\textit{h}}_t}$, ${\textbf{\textit{h}}_v}$, and ${\textbf{\textit{h}}_a}$). The \emph{Decoupling Module} decomposes ${\textbf{\textit{h}}_t}$, ${\textbf{\textit{h}}_v}$ and ${\textbf{\textit{h}}_a}$ into a modality-shared representation $\mathbf{S}$ and three modality-individual representations ${\textbf{\textit{i}}_t}$, ${\textbf{\textit{i}}_v}$, and ${\textbf{\textit{i}}_a}$. Finally, the \emph{Hierarchical Fusion Module} performs the fusion of these interaction representations.}
\label{fig:pipeline}
\end{figure*}

\section{METHOD}
This section describes our framework EffMulti for multimodal emotion analysis from the unimodal signals of text, audio and vision in a talking-face clip. Each clip in the data is segmented into constituent utterances bounded by breaths or pauses. Figure \ref{fig:pipeline} provides an overview of our framework. For simplicity, we define $\textbf{\textit{u}}_m^l$, $\textbf{\textit{h}}_m$ and $\textbf{\textit{i}}_m$ as the temporal utterances features, the utterance-based unimodal representation and the modality-individual representation (${m \in \{t, v, a \}}$ where $t$ refers to text, $a$ to audio and $v$ to vision; $l$ refers to temporal sequences of modality $m$.). 

Taking signals from three modalities as inputs, we propose three intermediate multimodal interaction representations to model complex multimodal interactions from different views. Then, these three intermediate representations are reasonably merged into a comprehensive interaction representation to unleash the full representation capacity of multimodal interaction. The framework consists of a unimodal encoding module, a decoupling operation and a hierarchical fusion module. More details will be described as follows.

\subsection{Unimodal encoding module}
\label{sec:uniModule}
Unimodal encoding module $\mathbb{E}_m$ (${{m} \in \{{t, v, a} \}}$) first computes utterance features  $\textbf{\textit{u}}_m^l$ from three-modal signals and then aligns them as utterance-based 64-dimensional unimodal representations  $\textbf{\textit{h}}_m$, respectively. The end-to-end trainable  $\textbf{\textit{h}}_m$ allows for reducing the modality gaps between heterogeneous signals and aligning them to the same representation space as much as possible, facilitating later on modeling their interactions. Then, three unimodal encoders $\mathbb{E}_m$ models the temporal relationships of utterances features $\textbf{\textit{u}}_m^l$ and transforms them into the utterance-based representations $\textbf{\textit{h}}_m$. 

\subsection{Decoupling representation}
As mentioned above, $\mathbb{E}_m$ aligns three-modal representations  $\textbf{\textit{h}}_m$. To model complex interactions, we develop an instance-based decoupling method to decompose  $\textbf{\textit{h}}_m$ into two decoupling representations, including the modality-shared representation $\mathcal{S}$ and three modality-individual representations $\textbf{\textit{i}}_t$/ $\textbf{\textit{i}}_a$/ $\textbf{\textit{i}}_v$. $\mathcal{S}$ is a common representation for three modalities. To facilitate training, the instance-based decoupling requires neither additional trainable parameters nor auxiliary constraint loss functions, as follows.

$\mathcal{S}$ is viewed as the average of  $\textbf{\textit{h}}_m$ from an instance. Then $\textbf{\textit{i}}_m$ is computed by subtracting $\mathcal{S}$ from  $\textbf{\textit{h}}_m$. The subtraction moves out the shared information from  $\textbf{\textit{h}}_m$ and reserve the individual information for each modality. In this way, we obtain three modality-individual representations with low similarity for each instance. The self-decoupling operation is formulated as follows:
\begin{equation}
\begin{array}{cc}
\mathcal{S} &= \frac{1}{3} \sum\limits_{m} {\textbf{\textit{h}}_m} ; \quad  {m \in \{{t, v, a} \}}  \\ 
\textbf{\textit{i}}_m &= \textbf{\textit{h}}_m - \mathcal{S}; \quad  {m \in \{{t, v, a} \}} \\
\end{array}
\end{equation}

To make modality-individual representations more efficient, a fully connected layer is introduced to individually transform each modality-individual representation $\textbf{\textit{i}}_m$ into the 16-dimensional representation $\tilde{\textbf{\textit{i}}}_m$. 


\subsection{Hierarchical fusion}
The hierarchical fusion aims at merging multiple representations ($\mathcal{S}$, $\tilde{\textbf{\textit{i}}}_m$ and  ${\textbf{\textit{h}}}_m$) for emotion analysis. According to the modality-semantic information, these representations are classified into three groups: the intact unimodal representations (${\textbf{\textit{h}}}_m$), the modality-individual representations ($\tilde{\textbf{\textit{i}}}_m$), and the modality-shared representation ($\mathcal{S}$).  

In the first step of hierarchical fusion, ${\textbf{\textit{h}}}_m$ are integrated into the interaction representation $\mathcal{M}$; $\tilde{\textbf{\textit{i}}}_m$ are integrated into the interaction representation $\mathcal{I}$. The underlying idea is that $\mathcal{M}$, $\mathcal{S}$, and $\mathcal{I}$ all have three-modal information but encode three-modal interactions from different views: $\mathcal{S}$ focuses on the multimodal coordination interaction for emotion analysis; $\mathcal{M}$ models three-modal interactions among their modality-intact representations; $\mathcal{I}$ emphasizes the three-modal modality-individual interaction. Such an encoding from multiple views allows for modeling sophisticated and different interactions.

Then, $\mathcal{S}$, $\mathcal{I}$ and $\mathcal{M}$ are concatenated together, denoted as comprehensive interaction representation $\mathcal{F}$. Finally, $\mathcal{F}$ is used for emotion analysis. More details can be seen below.

\begin{table*}[t]
\small
\centering
\footnotesize
\caption{Performances of emotion analysis on MOSEI and MOSI with EffMulti (Ours) and other baselines. $w/o$ $V$ refers to moving out the visual modality. In the column of Acc-2 and F1, The */* refers to two results on the metric of \emph{negative/non-negative} or \emph{negative/positive}, respectively. And $'-'$ denotes that the corresponding terms are not reported in the original works.}
\begin{tabular}{cc|c|c|c|c|c|c|c|c|c|cc}
\hline
\multicolumn{2}{c|}{\multirow{2}{*}{Method}} & \multicolumn{5}{c|}{MOSEI} & \multicolumn{5}{c}{MOSI} \\ \cline{3-12} 
\multicolumn{2}{c|}{} & \multicolumn{1}{c|}{MAE $\downarrow$} & \multicolumn{1}{c|}{Corr $\uparrow$} & \multicolumn{1}{c|}{Acc-2 $\uparrow$} & \multicolumn{1}{c|}{F1 $\uparrow$} & \multicolumn{1}{c|}{Acc-7 $\uparrow$} & \multicolumn{1}{c|}{MAE $\downarrow$} & \multicolumn{1}{c|}{Corr $\uparrow$} & \multicolumn{1}{c|}{Acc-2 $\uparrow$} & \multicolumn{1}{c|}{F1 $\uparrow$} & \multicolumn{1}{c}{Acc-7 $\uparrow$} \\ \hline
& MFN(C)              & -     & -      & 76.0/-    & 76.0/-   & -     & 0.965 & 0.632  & 77.4/-    & 77.3/-    & 34.1 \\
& MV-LSTM(C)          & -     & -      & 76.4/-    & 76.4/-   & -     & 1.019 & 0.601  & 73.9/-    & 74.0/-    & 33.2 \\
& MulT(C)  & 0.580 & 0.703 & -/82.5 & -/82.3 & 51.8 & 0.871 & 0.698 & -/83.0 & -/82.8 & 40.0 \\
& RAVEN(C)            & 0.614 & 0.662  & 79.1/-    & 79.5/-   & 50.0  & 0.915 & 0.691  & 78.0/-    & 76.6/-    & 33.2 \\
& \textbf{EffMulti}(C)     & \textbf{0.543} &\textbf{ 0.764}  & \textbf{83.4/85.7} &\textbf{83.6/85.8} & \textbf{51.8}  & \textbf{0.793} & \textbf{0.756}  & \textbf{81.1/83.0} &\textbf{80.8/82.0}  & \textbf{42.1} \\            \hline \hline
& LMF($B^{T}$)           & 0.623 & 0.677  & -/82.0    & -/82.1   & 48.0  & 0.917 & 0.695  & -/82.5    & -/82.4    & 33.2 \\
& TFN($B^{T}$)           & 0.593 & 0.700  & -/82.5    & -/82.1   & 50.2  & 0.901 & 0.698  & -/80.8    & -/80.7    & 34.9 \\ 
& MFM($B^{T}$)           & 0.568 & 0.717  & -/84.4    & -/84.3   & 51.3  & 0.877 & 0.706  & -/81.7    & -/81.6    & 35.4 \\ 
& ICCN($B^{T}$)          & 0.565 & 0.713   & -/84.2    & -/84.2   & 51.6  & 0.860 & 0.710  & -/83.0    & -/83.0    & 39.0 \\ 
& MISA($B^{T}$)          & 0.555 & 0.756  & 83.6/85.5   &83.8/85.3  & 52.2  & 0.783 & 0.761 & 81.8/83.4 & 81.7/83.6 & 42.3 \\
& \textbf{EffMulti}($B^{T}$)  &\textbf{0.541} &\textbf{0.765}   & \textbf{84.0/85.9} &\textbf{84.0/86.0} & \textbf{52.6}  &\textbf{0.761}  &\textbf{0.774}   &\textbf{81.8/83.8}  &\textbf{81.7/83.8}  & \textbf{45.1}  \\             \hline \hline

& SSL($B^{TS}$, $w/o$ $V$) & 0.491 & -    & -/88.0    & -/88.1    & 56.0 &\textbf{0.577}  & -     & -/88.3    & -/88.6   & 47.2\\ 
& \textbf{EffMulti} ($B^{TS}$, $w/o$ $V$)  & \textbf{0.483} & \textbf{0.773} & \textbf{87.0/88.3} & \textbf{86.9/88.1} & \textbf{56.5} & 0.586 & \textbf{0.871} & \textbf{87.8/89.2} & \textbf{87.8/89.1} & \textbf{51.5} \\ \hline

\end{tabular}
\label{table:moseimosi}
\end{table*}

\textbf{High-order representation.} 
To fully model complex interaction among modality-specific representations (${\textbf{\textit{h}}}_m$ or $\tilde{\textbf{\textit{i}}}_m$), the outer product is utilized to integrated them, allowing for learning their high-order correlations \cite{he2018outer}. It tightly entangles the learning process of multimodal representation elements with a multiplicative interaction between all elements among multimodal representations. Therefore, it avoids modality-bias overfitting and has been proved beneficial to the fusion of multimodal features.

Specifically, to avoid the curse of dimension, we first reduce the dimension of  $\textbf{\textit{h}}_m$, denoted as $\tilde{\textbf{\textit{h}}}_m$, from 64 to 16 through a fully-connected layer. Then $\tilde{\textbf{\textit{h}}}_m$ are merged into a high-order tensor (16$\times$16$\times$16) through the outer product. Additionally, to further reduce the trainable parameters, a Max Pooling Layer is introduced to reduce the high-order tensor into 8$\times$8$\times$8. Furthermore, the tensor is flattened and reduced into a 16-dimensional integrating representation $\mathcal{M}$ with a fully connected layer, which is formulated as follows:
\begin{equation}
  \mathcal{M} = \textit{FC} (\textit{Pool}(\tilde{\textbf{\textit{h}}}_t \otimes \tilde{\textbf{\textit{h}}}_v \otimes \tilde{\textbf{\textit{h}}}_a )) \\
\end{equation}


Similarly, three modality-individual representations $\tilde{\textbf{\textit{i}}}_t$ are merged into the integrating modality-individual representation $\mathcal{I}$. The merging is formulated as follows:
\begin{equation}
  \mathcal{I} = \textit{FC} (\textit{Pool}(\tilde{\textbf{\textit{i}}}_t \otimes \tilde{\textbf{\textit{i}}}_v \otimes \tilde{\textbf{\textit{i}}}_a )) \\
\end{equation}

\textbf{Fusion representation}. Three intermediate interaction representations $\mathcal{S}$, $\mathcal{I}$ and $\mathcal{M}$ are then concatenated into a comprehensive interaction representation $\mathcal{F}_0$. Finally, $\mathcal{F}$ is fed into three fully connected layers for emotion analysis. 





\section{EXPERIMENT}
Extensive experiments are conducted to evaluate the proposed EffMulti by comparing to baselines and by analysis on its performance. Moreover, ablation studies are conducted to look into the performance of the EffMulti. 

\textbf{Implementation details.} The optimization uses the Adam optimizer with learning rate 0.0001. The training duration of each model is governed by an early-stopping strategy with the patience of 10 epochs. The size of the mini-batch is 64. The training relies on one GTX3090 GPU. The partition of the data for training, validation, and testing follows the official setting, which is adopted by most of the previous works. The standard cross-entropy loss is employed. The codes will be released upon acceptance.

\textbf{Datasets.}
\label{secData}
Our experiments perform on the mentioned-before widely-used MOSI \cite{zadeh2016multimodal} and MOSEI \cite{zadeh2018multimodal}. 
MOSI contains 2,199 utterance samples from YouTube, spanning 89 different speakers. Each sample is manually annotated with seven emotional intensities ranging from $-3$ to $+3$, where -3/+3 indicates a strong negative/positive emotion. MOSEI is an extended version of MOSI with unified annotation labels. It includes 23,453 annotated utterance samples, 1,000 distinct speakers, and 250 different topics. 


\textbf{Preprocessing.} Traditionally, text modality features are GloVe \cite{pennington2014glove} embeddings; for visual modality, previous works use Facet to extract facial expression features \cite{ekman1980facial}; for acoustic modality, COVAREP \cite{degottex2014covarep} is used to extract the features. We call these three features as the \textit{classical} features denoted as $C$. Recent work \cite{MISA} have demonstrated that BERT \cite{devlin2018bert} can provide better features than GloVe for emotion analysis and boost the performance. Later on, another work \cite{twobert} further introduces Speech-BERT \cite{speechBert} to extract audio representations. For a fair comparison, we also conduct experiments on BERT features and Speech-BERT features denoted as $B^T$ and $B^S$, respectively. 
Additionally, we perform word-level alignment to obtain aligned data by the prevalent tool kit P2FA, which is regularly employed in previous works.

\subsection{Comparisons with baselines}
\label{sec:exp}
The baseline methods include MFN \cite{MFN}, MV-LSTM \cite{MVLSTM}, RAVEN \cite{wordsCShift}, MulT \cite{MulT}, TFN \cite{zadeh2017tensor}, LMF \cite{LMF}, MFM \cite{tsai2018learning}, ICCN \cite{ICCN}, MISA \cite{MISA} and SSL\cite{twobert}. As done in their works, MAE (mean absolute error), Corr (Pearson correlation coefficient), Acc-7 (7-class accuracy), Acc-2 (binary accuracy) and F1 score are taken as the metrics. Tables \ref{table:moseimosi} show the results. The partition of the data for training, validation, and testing follows the official setting adopted by previous works.

\textbf{Results.}
To make fair comparisons with these baselines, the results with our method rely on the identical features with the baselines. As can be seen in Table \ref{table:moseimosi}, our EffMulti achieves the best performance on most metrics. In comparison with those methods followed with ($C$) using the classical features, our results outperform all the baselines on MOSI and MOSEI. Specifically, our EffMulti outperforms MulT on all the metrics on MOSI and MOSEI. Looking into those methods using the classical audio and vision features but text-Bert features($B^{T}$), our EffMulti outperforms all the baselines. Particularly, comparing to SSL using two-modal features of text-Bert and audio-Bert, our EffMulti ($B^{TS}$) performs better on nine metrics, except for MAE on the smaller dataset of MOSI. The above observations validate our EffMulti.  

\subsection{Ablation study}
\label{sec:ab}

For simplicity, Table \ref{table:Abla} reports the ablation study results relying on the classical features on MOSEI and MOSI, and we only report on the metric of \emph{negative/non-negative} on F1. 

\begin{table}[tbp]
\setlength{\tabcolsep}{4pt}
\centering
\footnotesize
\caption{Results of the ablation study. All the experiments employ the classical features.} 
\begin{tabular}{|c|c|c|c|c|c|c|c|}
\hline
\multicolumn{2}{|c|}{Dataset} & \multicolumn{3}{c|}{MOSEI} & \multicolumn{3}{c|}{MOSI}  \\ \hline
\multicolumn{2}{|c|}{\multirow{1}{*}{Ablation Study}} & \multirow{1}{*}{MAE $\downarrow$} & \multirow{1}{*}{Corr $\uparrow$} & \multirow{1}{*}{F1 $\uparrow$} & \multirow{1}{*}{MAE $\downarrow$} & \multirow{1}{*}{Corr $\uparrow$} & \multirow{1}{*}{F1 $\uparrow$} \\\hline
\textbf{A0} & \textbf{EffMulti}    & \textbf{0.543} & \textbf{0.764} & \textbf{85.8} & \textbf{0.793} & \textbf{0.756} & \textbf{82.0} \\ \hline
A1 & $\mathcal{S}$ + $\mathcal{M}$             & 0.549 & 0.759 & 83.2 & 0.807 & 0.740 & 81.2 \\
A2 & $\mathcal{S}$ + $\mathcal{I}$               & 0.546 & 0.764 & 85.4 & 0.819 & 0.735 & 79.3 \\
A3 & $\mathcal{M}$   + $\mathcal{I}$               & 0.547 & 0.760 & 85.4 & 0.830 & 0.726 & 81.3 \\
A4 & $\mathcal{S}$ only            & 0.546 & 0.759 & 84.8 & 0.841 & 0.717 & 79.6 \\
A5 & $\mathcal{M}$   only            & 0.572 & 0.740 & 85.0 & 0.831 & 0.739 & 81.3 \\
A6 & $\mathcal{I}$   only            & 0.545 & 0.758 & 85.0 & 0.811 & 0.741 & 80.9  \\ \hline
A7 & w/o text          & 0.823 & 0.273 & 67.7 & 1.455 & 0.015 & 59.4 \\ 
A8 & w/o visual        & 0.551 & 0.755 & 85.5 & 0.880 & 0.708 & 80.6 \\ 
A9 & w/o acoustic      & 0.553 & 0.764 & 85.0 & 0.873 & 0.719 & 81.7 \\ \hline 
A10 & ortho. constraint & 0.557 & 0.757 & 85.5 & 0.813 & 0.744 & 82.0  \\ \hline
\end{tabular}
\label{table:Abla}
\end{table}

To confirm the necessity of three interaction representations in our proposed framework, we conduct ablation studies A1-A6 in Table \ref{table:Abla}. A1-A6 are conducted by employing one or two representations of {$\mathcal{S}$, $\mathcal{I}$ and $\mathcal{M}$} while keeping other parts as the same as A0. The results validate the effectiveness of our proposed framework. The proposed method (A0) with $\mathcal{S, M, I}$ is better than using one or two of them. It proves that these three interaction representations carry on different interaction information and are capable of modeling the complex multimodal interactions together. All of $\mathcal{S, M, I}$ cannot be neglected and cannot replace each other.

To confirm the capability of fitting to each modality, we conduct A7-A9 in Table \ref{table:Abla} by using only two modalities information as inputs and ignoring one modality of text, audio or vision, respectively. It is found that A7-A9 are worse than EffMulti (A0) using all three modalities. It means that EffMulti is capable of extracting meaningful but different information from each modality. 

To furthermore confirm our decoupling operation, we conduct A10 which replaces self-decoupling operation in EffMulti (A0) by the orthogonal constraints (including the similarity, difference and reconstruction losses) inspired by MISA \cite{MISA}. The result shows that the performance of A10 goes worse in comparison with A0. This observation furthermore validates our self-decoupling operation in the end-to-end training.

All the above experimental results confirm the effectiveness of EffMulti from three-modal signals, the use of three interaction representations $\mathcal{S, M, I}$, the developed decoupling mechanism. 

\section{CONCLUSION}
In this paper, we propose a novel carefully-designed deep learning method (EffMulti) for multimodal emotion analysis. The experiments validate the effectiveness of modeling the complex multimodal interactions from different views. The EffMulti outperforms the state-of-the-art methods on most metrics and demonstrates its compelling performance on the widely-used datasets. More than that, extensive experiments demonstrate the advance of the proposed method in better performance, ease of training, fast convergence, lower computing complexity, and less trainable parameters. 

A novel decoupling operation is developed to learn modality-shared and modality-individual interaction representations. It benefits from ease of implementation, fewer additional trainable parameters and no additional complex constraints. The end-to-end training allows for learning appropriate representation spaces for the above interaction representations. Based on that, a powerful hierarchical fusion is proposed to unleash the full representation capacity of the multimodal interaction for emotion analysis. 

In the future, we intend to extend EffMulti with intra-utterance fusion and try to achieve better performance. Additionally, we plan to dig deeper to the similarity of different features and further reduce the gap of different modalities.

\bibliographystyle{IEEEbib}
\bibliography{main}
\end{document}